\documentclass[11pt]{article}

\usepackage[margin=1.2in]{geometry}
\usepackage[utf8]{inputenc}
\usepackage[T1]{fontenc}
\usepackage{microtype}
\usepackage{graphicx}
\usepackage{booktabs}
\usepackage[numbers,sort&compress]{natbib} 
\usepackage{hyperref}
\usepackage{authblk} 
\usepackage{enumitem} 

\usepackage{tikz}
\usetikzlibrary{matrix, positioning, arrows.meta, calc, shapes.geometric}

\usepackage{amsmath}
\usepackage{amssymb}
\usepackage{mathtools}
\usepackage{amsthm}

\theoremstyle{plain}
\newtheorem{theorem}{Theorem}[section]

\theoremstyle{definition}
\newtheorem{definition}[theorem]{Definition}

\theoremstyle{remark}
\newtheorem{remark}[theorem]{Remark}

\newcommand{\vct}[1]{\mathbf{#1}}

\newcommand{\Wten}{\mathcal{W}} 
\newcommand{\R}{\mathbb{R}}

\title{How Many Heads Make an SSM?\\
\large A Unified Framework for Attention and State Space Models}

\author{Ali Ghodsi}
\affil{University of Waterloo, Canada \\ \texttt{ali.ghodsi@uwaterloo.ca}}

\date{}

\begin{document}

\maketitle

\begin{abstract}
Sequence modeling has produced diverse architectures---from classical recurrent neural networks to modern Transformers
and state space models (SSMs)---yet a unified theoretical understanding of expressivity and trainability trade-offs
remains limited. We introduce a \textbf{unified framework} that represents a broad class of sequence maps via an
input-dependent \emph{effective interaction operator} $\Wten_{ij}(X)$, making explicit two recurring construction
patterns: (i) the \textbf{Unified Factorized Framework (Explicit)} (attention-style mixing), in which
$\Wten_{ij}(X)$ varies through scalar coefficients applied to shared value maps, and (ii) \textbf{Structured Dynamics (Implicit)} (state-space recurrences), in which $\Wten_{ij}$ is induced by a latent dynamical system.

Using this framework, we derive three theoretical results. First, we establish the \textbf{Interaction Rank Gap}:
models in the Unified Factorized Framework, such as single-head attention, are constrained to a low-dimensional
operator span and cannot represent certain structured dynamical maps. Second, we prove an \textbf{Equivalence
(Head-Count) Theorem} showing that, within our multi-head factorized class, representing a linear SSM whose lag
operators span a $k$-dimensional subspace on length-$n$ sequences requires and is achievable with $H=k$ heads. Third,
we prove a \textbf{Gradient Highway Result}, showing that attention layers admit inputs with distance-independent gradient paths,
whereas stable linear dynamics exhibit distance-dependent gradient attenuation. Together, these results formalize a
fundamental trade-off between algebraic expressivity (interaction/operator span) and long-range gradient propagation,
providing theoretical grounding for modern sequence architecture design.
\end{abstract}

\section{Introduction: The Landscape of Explicit and Implicit Sequence Models}

The field of sequence modeling has arguably bifurcated into two dominant paradigms. On one hand, modern architectures
like the Transformer \citep{vaswani2017attention} and its variants rely on \emph{explicit} token-to-token interactions.
On the other hand, classical and recent models like Recurrent Neural Networks (RNNs) \citep{elman1990finding} and State
Space Models (SSMs) \citep{gu2022efficiently, gu2023mamba} represent an \emph{implicit} paradigm through recurrent
dynamics and hidden states. Despite their empirical success, these architectures are often studied in isolation, with
distinct theoretical vocabularies---geometry and kernels for Attention, versus control theory and differential
equations for SSMs. This separation obscures fundamental questions: Are ``multi-head'' attention mechanisms merely an
ensemble technique, or do they serve a necessary algebraic function? Why do input-dependent weighting schemes
(Attention) often train more robustly over long contexts than input-independent recurrent transition dynamics, even
when the latter can be highly expressive?

Convolutional Neural Networks (CNNs) and Kolmogorov--Arnold Networks (KANs)~\citep{liu2024kan} further broaden the
design space. While CNNs implement input-independent, translation-structured mixing, KAN-style constructions occupy an
intermediate regime in which the effective weights depend on \emph{one side} of the interaction (e.g., the source
token), whereas attention depends on \emph{both} the source and target tokens. This observation motivates treating a
wide range of architectures---including convolutional and feed-forward models, which are not inherently sequential or
stateful but are widely applied to sequential data---within a single representational lens.

In this work, we propose a \textbf{Unified Framework} to bridge these gaps. We show that many commonly used
architectures---MLPs, attention variants, and linear state-space models (and certain restricted/separable
convolutions)---can be cast as different constructions of an input-dependent effective weight representation
$\Wten(X)$. This unification allows us to move beyond descriptive comparisons and derive rigorous results. Specifically,
we identify an \textbf{Interaction Rank Gap}, showing that scalar-factorized interactions (e.g., single-head
attention-style mixing) are algebraically rank-limited relative to the interaction families induced by structured
dynamics, and that multiple heads are required to match higher-rank interaction subspaces in our setting.

\section{Related Work: Efficient Attention and Implicit Dynamics}

The quest to optimize sequence modeling has largely followed two complementary trajectories: reducing the cost of
explicit attention and improving the expressivity of implicit recurrent dynamics.

\textbf{Efficient Attention Mechanisms.}
Standard attention forms query--key interactions across all pairs of positions, incurring $O(n^2)$ time and memory for
sequence length $n$. To mitigate this cost, kernel-based methods such as Linear Attention
\citep{katharopoulos2020transformers} and Performer \citep{choromanski2021performer} approximate the softmax attention
kernel using low-rank feature maps, enabling $O(n)$ computation. These methods can also be interpreted in recurrent
form by reordering the computation and maintaining suitable running sums \citep{katharopoulos2020transformers}. While
effective, such approximations introduce an accuracy--efficiency trade-off relative to full softmax attention in some
settings. Our framework provides a complementary perspective by characterizing limitations that arise from
scalar-factorized interaction parameterizations via interaction-rank constraints.

\textbf{Structured State Space Models.}
Conversely, SSM-based architectures aim to endow recurrent models with strong long-range modeling capacity while
preserving linear-time computation. The Structured State Space (S4) line
\citep{gu2022efficiently} introduces structured transitions that enable efficient
computation of long convolution kernels. More recently, Mamba \citep{gu2023mamba} incorporates input-dependent
(\emph{selective}) mechanisms into the SSM update, moving beyond purely input-independent transition dynamics. Our work
formalizes a unifying lens in which both fixed and input-dependent interaction rules can be expressed through an
effective weight representation, and it clarifies how input dependence can create qualitatively different long-range
sensitivity behavior.

\textbf{Unification Efforts.}
Prior work has drawn connections between specific architectures, most notably the ``Transformers are RNNs''
perspective \citep{katharopoulos2020transformers}, which links linear attention to recurrent computations. However,
many existing unifications focus on expressing one model class as an efficient approximation of another. In contrast,
our work provides a common representational framework that encompasses attention mechanisms and linear state-space
models (and, in additional sections, convolutional and feed-forward constructions), enabling rigorous separation and
equivalence results (Theorems~\ref{thm:rank_gap_attn} and \ref{thm:equivalence}) about the expressivity of
scalar-factorized interactions versus structured dynamical interactions.

\section{The Unified  Framework}
\label{sec:framework}

We define a sequence model as a transformation mapping an input sequence $X = [\vct{x}_1, \dots, \vct{x}_n] \in \R^{d \times n}$ to an output sequence $Y = [\vct{y}_1, \dots, \vct{y}_n] \in \R^{p \times n}$. While traditional feedforward networks employ fixed weights (e.g., $Y=VX$), modern sequence models require weights that adapt to the context.

We propose that all such models can be unified under a single formulation where the transformation is governed by an input-dependent weight tensor $\Wten(X) \in \R^{n \times n \times p \times d}$. The output token $\vct{y}_i$ is computed as a weighted sum of all input tokens $\vct{x}_j$, mediated by the sub-tensor $\Wten_{ij} \in \R^{p \times d}$ which represents the specific linear map between position $j$ and position $i$:
\begin{equation}
    \vct{y}_i = \sum_{j=1}^n \Wten_{ij}(X) \vct{x}_j
    \label{eq:unified_general}
\end{equation}
The tensor $\Wten_{ij}$ captures the effective weight matrix applied to token $\vct{x}_j$ to produce its contribution to $\vct{y}_i$. Direct instantiation of this 4D tensor requires $O(n^2 pd)$ parameters, which is computationally intractable. Consequently, all practical architectures can be viewed as distinct strategies for factorizing or implicitly defining $\Wten_{ij}$.

\begin{figure}[ht]
    \centering
    \begin{tikzpicture}[
        scale=0.6,
        matrix_node/.style={draw, rectangle, minimum size=0.4cm, anchor=center},
        label_node/.style={font=\bfseries\small}
    ]
        \node[label_node] at (2, 5) {MLP};
        \node[label_node] at (2, 4.5) {(Identity)};
        \draw[step=0.5cm, gray!20, thin] (0,0) grid (4,4);
        \draw[thick] (0,0) rectangle (4,4);
        \foreach \i in {0.25, 0.75, ..., 3.75}
            \fill[black] (\i, 4-\i) circle (0.15cm);

        \begin{scope}[xshift=6cm]
            \node[label_node] at (2, 5) {CNN};
            \node[label_node] at (2, 4.5) {(Banded/Toeplitz)};
            \draw[step=0.5cm, gray!20, thin] (0,0) grid (4,4);
            \draw[thick] (0,0) rectangle (4,4);
            \foreach \i in {0.25, 0.75, ..., 3.75} {
                \fill[black] (\i, 4-\i) circle (0.15cm);
                \ifdim \i pt > 0.5pt \fill[gray] (\i-0.5, 4-\i) circle (0.1cm); \fi
                \ifdim \i pt < 3.5pt \fill[gray] (\i+0.5, 4-\i) circle (0.1cm); \fi
            }
        \end{scope}

        \begin{scope}[xshift=12cm]
            \node[label_node] at (2, 5) {RNN / SSM};
            \node[label_node] at (2, 4.5) {(Causal History)};
            \draw[step=0.5cm, gray!20, thin] (0,0) grid (4,4);
            \draw[thick] (0,0) rectangle (4,4);
            \foreach \x in {0, ..., 7} {
                \foreach \y in {0, ..., \x} {
                    \fill[black!70] ({(\y+0.5)/2}, {4-(\x+0.5)/2}) circle (0.1cm);
                }
            }
        \end{scope}

        \begin{scope}[xshift=18cm]
            \node[label_node] at (2, 5) {Attention};
            \node[label_node] at (2, 4.5) {(Input-Dependent)};
            \draw[step=0.5cm, gray!20, thin] (0,0) grid (4,4);
            \draw[thick] (0,0) rectangle (4,4);
             \foreach \x in {0, ..., 7} {
                \foreach \y in {0, ..., 7} {
                    \pgfmathsetmacro{\op}{rnd}
                    \ifdim \op pt > 0.3pt
                        \fill[blue, opacity=\op] ({(\y+0.5)/2}, {4-(\x+0.5)/2}) circle (0.12cm);
                    \fi
                }
            }
        \end{scope}
    \end{tikzpicture}
    \caption{\textbf{Visualizing the Interaction Structure.} The effective scalar weight matrix $A(X)$ (where $Y=VXA^\top$) for different architectures. While MLPs and CNNs enforce static sparsity patterns, and RNNs enforce triangular causality, Attention generates a dense, input-dependent interaction graph.}
    \label{fig:interaction_patterns}
\end{figure}
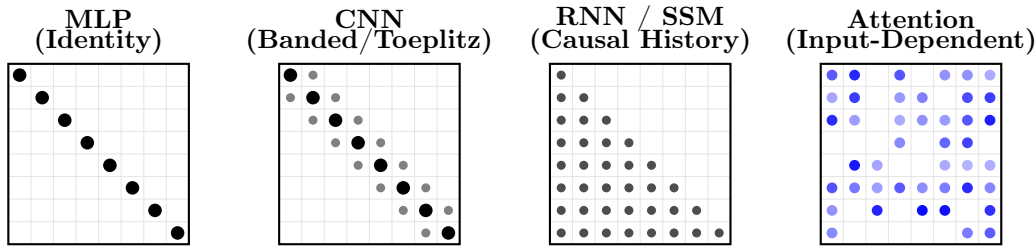

\subsection{Strategy I: The Unified Factorized Framework (Explicit)}
The most common strategy is to constrain the interaction matrix $\Wten_{ij}$ to be a rank-1 factorization of a scalar interaction score $w_{ij}$ and a shared value matrix $V \in \R^{p \times d}$. This dramatically reduces the parameter count from $O(n^2 pd)$ to $O(n^2 + pd)$.

\begin{definition}[Unified Factorized Model]
A model belongs to the Unified Factorized class if its weight tensor decomposes as
\[
\Wten_{ij}(X)=f_{\theta}(\vct{x}_i,\vct{x}_j)\,V,
\]
where $V\in\R^{p\times d}$ is a shared value matrix and $f_{\theta}:\R^d\times\R^d\to\R$ is a scalar token-pair weight function. The output equation becomes
\[
\vct{y}_i=\sum_{j=1}^{n} f_{\theta}(\vct{x}_i,\vct{x}_j)\,(V\vct{x}_j).
\]
In matrix notation, defining the scalar weight matrix $A(X) \in \R^{n \times n}$ with entries
$A_{ij}=f_{\theta}(\vct{x}_i,\vct{x}_j)$, the operation factorizes as
\[
Y = V X A(X)^\top.
\]
This highlights that the output is a linear combination of the transformed inputs $VX$.
In general, $\vct{x}_i$ may be understood to include any fixed positional features; thus $f_{\theta}$ can represent purely positional weights (e.g., $\delta_{ij}$ or $c_{i-j}$) as well as content-dependent weights.
\end{definition}

Within this class, we distinguish between \textit{static} factorization, where $f_\theta$ depends only on positional indices, and \textit{dynamic} factorization, where $f_\theta$ depends on the content of $X$.

\subsubsection{Static Factorization (Fixed Weights)}
In these models, the interaction strength $w_{ij}$ is determined solely by the relative or absolute positions of tokens $i$ and $j$, independent of the token content.
\begin{itemize}
    \item \textbf{MLP (Feedforward):} The Multi-Layer Perceptron treats tokens independently. The interaction function is the Kronecker delta $f_\theta(\vct{x}_i, \vct{x}_j) = \delta_{ij}$. Substituting this into the unified equation yields $\vct{y}_i = \sum_j \delta_{ij} V \vct{x}_j = V \vct{x}_i$, which recovers the standard layer-wise transformation. Matrix Form: $Y = VX I = VX$. The effective weight interaction matrix is the Identity.

    \item \textbf{CNN (Convolution):}
    Local, shift-invariant weights: $f_\theta(\vct{x}_i, \vct{x}_j) = c_{i-j}$ for $|i - j| \le r$. Substituting this into Eq. (\ref{eq:unified_general}) gives:
    \[ \vct{y}_i = \sum_{|i-j|\le r} c_{i-j} V\vct{x}_j \]
    Change variable $k = i - j$, so $j = i - k$ and $|k| \le r$:
    \[ \vct{y}_i = \sum_{k=-r}^{r} c_k (V\vct{x}_{i-k}) \]
    This matches the definition of a 1D discrete convolution of the sequence $(VX)$ with kernel $[c_{-r}, \dots, c_0, \dots, c_r]$.
    Matrix Form: $Y = VX C_{\text{Toeplitz}}^\top$, where $C_{\text{Toeplitz}}$ is a banded matrix reflecting the local, shift-invariant kernel structure.
\end{itemize}

\subsubsection{Dynamic Factorization (Input-Dependent Weights)}
These models allow the weight tensor $\Wten(X)$ to adapt to the input sequence, enabling context-aware processing.
\begin{itemize}
    \item \textbf{KAN (Kolmogorov-Arnold Network):}
    Based on the Kolmogorov–Arnold representation theorem \citep{kolmogorov1957superposition,arnold1957functions}, we can approximate a multivariate function using a superposition of univariate functions. For token interactions, we use a separable nonlinear basis expansion with $r$ basis functions:
    \[ g_m : \R^d \to \R, \quad h_m : \R^d \to \R, \quad m = 1, \dots, r \]
    Each basis pair $(g_m, h_m)$ captures a different mode of interaction between tokens. Define the token-pair weight as a sum of separable basis functions:
    \[ f_\theta(\vct{x}_i, \vct{x}_j) = \sum_{m=1}^r g_m(\vct{x}_i) h_m(\vct{x}_j) \]
    Substituting into the unified framework:
    \[ \vct{y}_i = \sum_{j=1}^n f_\theta(\vct{x}_i, \vct{x}_j) V\vct{x}_j = \sum_{j=1}^n \sum_{m=1}^r g_m(\vct{x}_i) h_m(\vct{x}_j) V\vct{x}_j \]
    Reorder the summation to separate token-dependent and sequence-dependent terms:
    \[ \vct{y}_i = \sum_{m=1}^r g_m(\vct{x}_i) \left( \sum_{j=1}^n h_m(\vct{x}_j) V\vct{x}_j \right) \]
    Precompute the $r$ content channels:
    \[ C_m = \sum_{j=1}^n h_m(\vct{x}_j) V\vct{x}_j \in \R^p, \quad m = 1, \dots, r \]
    Then:
    \[ \vct{y}_i = \sum_{m=1}^r g_m(\vct{x}_i) C_m \]
    Each output token is a weighted combination of $r$ content channels, with weights $g_m(\vct{x}_i)$ depending on the query token. Within our Unified Framework, KANs fall strictly under the Dynamic Factorized category, where the weight generation mechanism is defined by the separable basis functions.

    \item \textbf{Attention:}
    Standard attention relies \cite{vaswani2017attention} on computing a similarity score between queries and keys. Let $Q = W_Q X$, $K = W_K X$, and $V_{val} = W_V X$. The attention matrix is computed as $A = \text{softmax}(\frac{Q^\top K}{\sqrt{d_k}}) \in \R^{n \times n}$, where softmax is applied row-wise. The output is $Y = V_{val} A^\top$.

    To map this to our framework, define token-level queries $q_i = W_Q \vct{x}_i$ and keys $k_j = W_K \vct{x}_j$. Set the token-pair weight function:
    \[ f_\theta(\vct{x}_i, \vct{x}_j) = \text{softmax}_j \left( \frac{\vct{q}_i^\top \vct{k}_j}{\sqrt{d_k}} \right) \]
    With $A_{ij} = f_\theta(\vct{x}_i, \vct{x}_j)$, the Unified Factorized Framework gives $Y = V X A^\top$. Since $V$ is the shared matrix applied to inputs, identifying $V = W_V$ yields $V X = W_V X = V_{val}$, recovering the standard formula $Y = V_{val} A^\top$.
    Matrix Form: $Y = VX A^\top$, where $A$ is the dense, input-dependent softmax matrix.

    \item \textbf{Linear Attention:}
    In Linear Attention \cite{katharopoulos2020transformers, choromanski2021performer} the softmax is replaced with a kernel approximation using feature maps $\phi, \psi : \R^d \to \R^r$ where $r \ll n$. Let $Q = W_Q X$, $K = W_K X$, $V_{val} = W_V X$.
    Define feature-mapped queries and keys:
    \[ \Phi = [\phi(q_1), \dots, \phi(q_n)] \in \R^{r \times n}, \quad \Psi = [\psi(k_1), \dots, \psi(k_n)] \in \R^{r \times n} \]
    The output is $Y = V_{val} (\Phi^\top \Psi)^\top = V_{val} \Psi^\top \Phi$. This reduces complexity from $O(n^2)$ to $O(nr)$.

    In the Unified Factorized Framework, for each token pair, define the weight function as the inner product of feature maps:
    \[ f_\theta(\vct{x}_i, \vct{x}_j) = \phi(\vct{q}_i)^\top \psi(\vct{k}_j) = \phi(W_Q \vct{x}_i)^\top \psi(W_K \vct{x}_j) \]
    The output becomes:
    \[ \vct{y}_i = \sum_{j=1}^n f_\theta(\vct{x}_i, \vct{x}_j) V\vct{x}_j = \sum_{j=1}^n \phi(\vct{q}_i)^\top \psi(\vct{k}_j) V\vct{x}_j \]
    Matrix Form: With $\Phi$ and $\Psi$ as defined, the weight matrix is $W(X) = \Phi^\top \Psi$. From the Unified Framework $Y = VX W(X)^\top = VX (\Phi^\top \Psi)^\top = V_{val} \Psi^\top \Phi$. This matches exactly the standard linear attention formula.

    The key insight is to reorder the summation. Starting from $\vct{y}_i = \sum_{j=1}^n \phi(\vct{q}_i)^\top \psi(\vct{k}_j) V\vct{x}_j$, factor out $\phi(\vct{q}_i)$:
    \[ \vct{y}_i = \phi(\vct{q}_i)^\top \left( \sum_{j=1}^n \psi(\vct{k}_j) (V\vct{x}_j)^\top \right) \]
    Precompute the summary matrix $S = \sum_{j=1}^n \psi(\vct{k}_j) (V\vct{x}_j)^\top \in \R^{r \times p}$. Then $\vct{y}_i = \phi(\vct{q}_i)^\top S \in \R^p$ with cost $O(nr + rp)$ instead of $O(n^2p)$.
\end{itemize}

\subsection{Strategy II: Structured Dynamics (Implicit)}
The second strategy, employed by RNNs \cite{elman1990finding} and SSMs \cite{kalman1960new, gu2022efficiently}, avoids explicitly materializing the $O(n^2)$ interactions. Instead, it defines $\Wten_{ij}$ implicitly through the impulse response of a linear dynamical system.
\begin{definition}[Structured Dynamic Model]
A model belongs to the Structured Dynamic class if $\Wten_{ij}$ is defined by the state evolution of a dynamical system. In its most general form (covering Mamba and Time-Varying RNNs):
\[
    \Wten_{ij} = C_i \left( \prod_{k=j+1}^{i} \bar{A}_k(\vct{x}_k) \right) \bar{B}_j(\vct{x}_j) \quad \text{for } j \le i
\]
where $\bar{A}_k \in \R^{s \times s}$ is the (possibly input-dependent) state transition matrix, $\bar{B}_j \in \R^{s \times d}$ is the input projection, and $C_i \in \R^{p \times s}$ is the output projection.
\end{definition}
This formulation generalizes the standard time-invariant SSM (like S4 \cite{gu2022efficiently}) where matrices are constant: $\Wten_{ij} = C \bar{A}^{i-j} \bar{B}$. Unlike the factorized case $\Wten_{ij} = w_{ij}V$, this matrix is generated by the powers or products of $\bar{A}$ and is not generally rank-1, providing higher expressivity at the cost of being historically harder to optimize.

\paragraph{Implicit Weights in Nonlinear RNNs.}
While the linear formulation exactly covers SSMs and linearized RNNs, standard RNNs employ nonlinear activations (e.g., $h_t = \sigma(W h_{t-1} + U x_t)$). In our Unified Framework, the interaction tensor $\Wten_{ij}$ for such models is rigorously defined as the input-output Jacobian:
\[ \Wten_{ij} = \frac{\partial \vct{y}_i}{\partial \vct{x}_j} \]
This definition captures the local sensitivity of the output at step $i$ to the input at step $j$. Unlike the constant impulse response of linear systems, the Jacobian for nonlinear RNNs is input-dependent, varying across different sequences. However, unlike Attention where the dependency is explicit and parallelizable, the dependency here is implicit and must be computed sequentially.

\paragraph{Computational Duality of Linear SSMs.}
A critical distinction within the Structured Dynamic class is the impact of linearity on computation. Nonlinear RNNs allow for complex state transitions but enforce \textit{sequential} computation ($O(n)$ depth), preventing parallel training on GPUs. In contrast, the linearity of SSMs enables a computational duality: the recurrence $h_i = \bar{A}h_{i-1} + \bar{B}x_i$ can be unrolled into a convolution operation $\vct{y} = \vct{k} * \vct{x}$, where the kernel $\vct{k}$ is derived from the power series of $\bar{A}$. This allows SSMs to switch modes: using efficient sequential recurrence for inference ($O(1)$ per step) and highly parallel convolution via FFT for training ($O(\log n)$ depth). This duality solves the optimization bottleneck of traditional RNNs while maintaining their inference efficiency.

\begin{table}[h]
\centering
\caption{Comparison of representative architectures under the Unified Framework. \textbf{Complexity}: arithmetic cost for a length-$n$ forward pass (one layer). \textbf{Memory}: auxiliary memory to materialize/accumulate interaction weights (e.g., storing an $n\times n$ attention matrix in standard implementations; memory-efficient variants may reduce this). \textbf{Depth}: minimum number of sequential steps (parallelism bottleneck); for linear/associative scans this can be $O(\log n)$.}
\label{tab:framework_comparison}
\vspace{0.1in}
\resizebox{\textwidth}{!}{
\begin{tabular}{@{}lllllll@{}}
\toprule
\textbf{Model} & \textbf{Strategy} & \textbf{Interaction Tensor} $\Wten_{ij}$ & \textbf{Weight Type} & \textbf{Complexity} & \textbf{Memory} & \textbf{Depth} \\ \midrule
MLP & Factorized & $\delta_{ij} V$ & Static & $O(n)$ & $O(1)$ & $O(1)$ \\
CNN (separable) & Factorized & $c_{i-j} V$ & Static & $O(n)$ & $O(1)$ & $O(1)$ \\
Attention & Factorized & $\alpha_{ij}(X)\,V,\;\alpha_{ij}=\mathrm{softmax}(\vct{q}_i^\top \vct{k}_j)$ & Dynamic & $O(n^2)$ & $O(n^2)$ & $O(1)$ \\
Linear Attn & Factorized & $(\phi(\vct{x}_i)^\top \psi(\vct{x}_j))\,V$ & Dynamic & $O(n)$ & $O(1)$ & $O(\log n)$ \\
RNN / SSM (linear) & Structured & $C\,\bar{A}^{\,i-j}\bar{B}$ & Static & $O(n)$ & $O(1)$ & $O(\log n)$ \\
Mamba / selective SSM & Structured & $C_i(X)\Big(\prod_{k=j+1}^{i}\bar{A}_k(X)\Big)\bar{B}_j(X)$ & Dynamic & $O(n)$ & $O(1)$ & $O(\log n)$ \\
\bottomrule
\end{tabular}
}
\end{table}

\section{Theoretical Analysis: The Interaction Rank Gap}
\label{sec:theory}

The unified framework reveals a fundamental algebraic distinction between the two strategies: Factorized models restrict the interaction sub-tensor $\Wten_{ij}$ to be a scalar multiple of a shared matrix $V$, whereas Structured models allow $\Wten_{ij}$ to be a full-rank matrix derived from system dynamics. We now quantify this gap by establishing an impossibility theorem for factorized models.

\subsection{Interaction Rank}
We define the \textit{Interaction Rank} of a model as the dimension of the subspace spanned by its set of effective pairwise linear maps. To formalize the approximation gap, we measure the error between the sequence-to-sequence functions induced by the models.

\begin{definition}[Functional Approximation Error]
For two sequence models $F, f: \R^{d \times n} \to \R^{p \times n}$, we define the approximation error over the unit ball of inputs:
\begin{equation}
    \|F - f\|_\infty = \sup_{X: \|X\|_F \le 1} \|F(X) - f(X)\|_F
\end{equation}
\end{definition}

\begin{theorem}[Interaction Rank Gap for Single-Head Factorized Models]
\label{thm:rank_gap_attn}
\leavevmode\\
Fix $n\ge 2$ and define the uniform (supremum) error over the unit Frobenius ball by
\[
\|F-f\|_\infty \;:=\; \sup_{\|X\|_F\le 1}\,\|F(X)-f(X)\|_F .
\]
Let $\mathcal M_{\mathrm{Fact}}$ be the class of sequence maps $f:\R^{d\times n}\to \R^{p\times n}$
for which there exist a \emph{single shared matrix} $V\in\R^{p\times d}$ and scalar functions
$\alpha_{ij}:\R^{d\times n}\to\R$ such that, for all inputs $X=[x_1,\dots,x_n]$ and all $i$,

\begin{equation}\label{eq:fact-class}
\hat y_i \;=\; \sum_{j=1}^n \alpha_{ij}(X)\,Vx_j.
\end{equation}

(This includes single-head softmax attention and single-head linear attention as special cases.)

Let $\mathcal M_{\mathrm{Dyn}}$ be the class of linear time-invariant SSM maps $F$ of the form
\[
y_i \;=\; \sum_{j=1}^{i} W(i-j)\,x_j,
\qquad W(\tau)=C\bar A^{\tau}\bar B.
\]

Then there exists $F\in\mathcal M_{\mathrm{Dyn}}$ whose impulse-response operators
$\{W(\tau)\}_{\tau=0}^{n-1}$ are not all collinear such that
\[
\inf_{f\in\mathcal M_{\mathrm{Fact}}}\|F-f\|_\infty \;\ge\; 1,
\]
independent of $n$. In particular, no single-head factorized model with one shared matrix $V$
can uniformly approximate $F$.

\begin{proof}
We give an explicit $F\in\mathcal M_{\mathrm{Dyn}}$ and lower bound $\|F-f\|_\infty$ for every
$f\in\mathcal M_{\mathrm{Fact}}$.

\par\medskip\noindent\textbf{Step 1: Define the target SSM.}
Take $d=p=2$, $\bar B=I_2$, $C=I_2$, and
\[
\bar A := \begin{pmatrix}0 & -1\\ 1 & 0\end{pmatrix},
\]
the $90^\circ$ rotation. Then $W(0)=I_2$ and $W(1)=\bar A$.
For a length-$n$ sequence $X=[x_1,\dots,x_n]$, the SSM $F$ produces
\[
y_1 = W(0)x_1 = x_1,\qquad
y_2 = W(1)x_1 + W(0)x_2 = \bar A x_1 + x_2,
\]
(and subsequent outputs are defined similarly, but will not be needed).

\par\medskip\noindent\textbf{Step 2: Restrict to a one-token input.}
Fix the input sequence
\[
X = [e_1,0,0,\dots,0]\in\R^{2\times n},
\]
where $e_1=(1,0)^\top$ and  $0\in\R^2$ denotes the zero vector.
 Then $\|X\|_F=\|e_1\|_2=1$.
For this $X$, the target outputs at times $1$ and $2$ are
\[
y_1 = e_1,\qquad y_2 = \bar A e_1 = e_2,
\]
where $e_2=(0,1)^\top$.

\par\medskip\noindent\textbf{Step 3: Form of any single-head factorized model on this input.}
Let $f\in\mathcal M_{\mathrm{Fact}}$ be arbitrary. By \eqref{eq:fact-class} and since $x_j=0$ for $j\ge 2$,
the first two outputs of $f$ on this $X$ have the form
\[
\hat y_1 = \alpha_{11}(X)\,V e_1,\qquad
\hat y_2 = \alpha_{21}(X)\,V e_1.
\]
Define $v := V e_1\in\R^2$ and scalars $a:=\alpha_{11}(X)$, $b:=\alpha_{21}(X)$. Then
\[
\hat y_1 = a v,\qquad \hat y_2 = b v.
\]
Therefore the squared error on the first two time steps equals
\begin{equation}\label{eq:err-two}
\|y_1-\hat y_1\|_2^2+\|y_2-\hat y_2\|_2^2
= \|e_1-av\|_2^2 + \|e_2-bv\|_2^2.
\end{equation}

\par\medskip\noindent\textbf{Step 4: A geometric lower bound independent of $v$.}

If $v=0$, then \eqref{eq:err-two} equals $\|e_1\|_2^2+\|e_2\|_2^2=2$ for all $a,b$.
In particular, $\|F-f\|_\infty \ge 2 \ge 1$.

Assume now $v\neq 0$ and let $L:=\mathrm{span}\{v\}$.
For fixed $v$, minimizing \eqref{eq:err-two} over scalars $a$ and $b$ is exactly projecting $e_1$ and $e_2$
onto the line $L$. Thus
\[
\inf_{a\in\R}\|e_1-av\|_2^2 = \mathrm{dist}(e_1,L)^2,\qquad
\inf_{b\in\R}\|e_2-bv\|_2^2 = \mathrm{dist}(e_2,L)^2.
\] where $\mathrm{dist}(u,L):=\inf_{z\in L}\|u-z\|_2$ denotes the Euclidean distance from $u$ to the subspace $L$.

Let $P_L$ be the orthogonal projector onto $L$. Then
\[
\mathrm{dist}(e_i,L)^2 = \|(I-P_L)e_i\|_2^2 = 1-\|P_L e_i\|_2^2,\qquad i\in\{1,2\}.
\]
Summing over $i=1,2$ gives
\[
\mathrm{dist}(e_1,L)^2+\mathrm{dist}(e_2,L)^2
= 2 - \big(\|P_L e_1\|_2^2+\|P_L e_2\|_2^2\big)
= 2 - \mathrm{tr}(P_L).
\]
Since $P_L$ is a rank-one orthogonal projector, $\mathrm{tr}(P_L)=1$. Hence
\begin{equation}\label{eq:proj-identity}
\inf_{a,b\in\R}\Big(\|e_1-av\|_2^2 + \|e_2-bv\|_2^2\Big) \;=\; 1.
\end{equation}

\par\medskip\noindent\textbf{Step 5: Conclude a uniform separation.}
By \eqref{eq:err-two}--\eqref{eq:proj-identity}, for our fixed feasible $X$,
\[
\|F(X)-f(X)\|_F^2 \;\ge\; \|y_1-\hat y_1\|_2^2+\|y_2-\hat y_2\|_2^2 \;\ge\; 1,
\]
so $\|F(X)-f(X)\|_F\ge 1$. Since $\|X\|_F=1$, this implies
\[
\|F-f\|_\infty
=\sup_{\|Z\|_F\le 1}\|F(Z)-f(Z)\|_F
\;\ge\;\|F(X)-f(X)\|_F
\;\ge\; 1.
\]
Because $f\in\mathcal M_{\mathrm{Fact}}$ was arbitrary, we obtain
$\inf_{f\in\mathcal M_{\mathrm{Fact}}}\|F-f\|_\infty\ge 1$, completing the proof.
\end{proof}
\end{theorem}

\begin{remark}
Although the proof is presented for $d=2$, the result extends to any $d\ge 2$ by embedding the same
two-dimensional construction into $\R^d$ (e.g., applying the $2\times 2$ rotation on the first two coordinates
and leaving the remaining coordinates unchanged). The impossibility for single-head factorized models already
follows from the existence of two non-collinear impulse-response operators.
\end{remark}

\subsection{Bridging the Gap: Multi-Head Equivalence}
Theorem~\ref{thm:rank_gap_attn} shows that a single-head factorized model with one shared value matrix $V$
can only generate interaction operators that are all scalar multiples of $V$. In other words, it can use
different \emph{weights} across pairs $(i,j)$, but it cannot produce genuinely different linear transforms at
different lags. A linear SSM, in contrast, can have impulse-response matrices $\Wten(\tau)$ that point in several
independent “directions” in operator space.

Multi-head factorization is exactly the mechanism that removes this bottleneck: with $H$ heads, the model can form
operators of the form $\sum_{h=1}^H \alpha^{(h)}_{ij} V^{(h)}$ \footnote{In standard multi-head attention, head outputs are concatenated and then multiplied by an output matrix
$W_O$.
This is equivalent to a sum-of-heads form after reparameterization: writing head $h$ as producing
$z_i^{(h)}=\sum_j \alpha^{(h)}_{ij}(X)\,V_h x_j$ and stacking $z_i=[z_i^{(1)};\dots;z_i^{(H)}]$, we have
$y_i=W_O z_i=\sum_{h=1}^H (W_O^{(h)} V_h)\sum_j \alpha^{(h)}_{ij}(X)\,x_j$,
where $W_O^{(h)}$ denotes the block of $W_O$ acting on head $h$.
Thus concatenation followed by $W_O$ can be absorbed into effective per-head matrices $V^{(h)}:=W_O^{(h)}V_h$.}
, i.e., it can mix up to $H$ independent matrices
$\{V^{(h)}\}$ and thereby span a higher-dimensional interaction subspace. The next theorem makes this precise for
linear SSMs by introducing the \emph{interaction rank}
$k=\dim(\mathrm{span}\{\Wten(\tau)\})$.
We show that to represent the SSM exactly on length-$n$ sequences, one needs at least $k$ heads ($H\ge k$), and
conversely $H\ge k$ heads are sufficient (with appropriate positional feature maps) to reproduce the same causal
input--output map. Thus, in this model class, the number of heads directly controls the dimension of the interaction subspace
that can be represented.

\begin{theorem}[Multi-Head Rank Constraint]\label{thm:equivalence}

\leavevmode\\
Let $\mathcal{M}_{\text{SSM}}$ be a linear state space model with impulse-response (interaction) matrices
\[
\Wten(\tau)\;=\;C\bar A^{\tau}\bar B\in\R^{p\times d},\qquad \tau\ge 0,
\]
and interaction rank
\[
k \;=\; \dim\Big(\mathrm{span}\{\Wten(\tau):\tau\ge 0\}\Big).
\]
Fix a sequence length $n$ and consider the causal input--output map \footnote{We impose causality to match the standard SSM input--output map. The same head-count argument applies
to the non-causal (bidirectional) setting by dropping the mask and summing over $j=1,\dots,n$.}

\[
\vct{y}_i \;=\; \sum_{j=1}^{i} \Wten(i-j)\,\vct{x}_j,\qquad i=1,\dots,n.
\]
A \emph{multi-head factorized linear-attention model} with $H$ heads, head dimension $r$, value matrices
$V^{(h)}\in\R^{p\times d}$, and positional feature maps
$\phi^{(h)},\psi^{(h)}:\{1,\dots,n\}\to\R^r$ produces outputs
\begin{equation}\label{eq:mh-linear-attn}
\hat{\vct{y}}_i \;=\; \sum_{j=1}^{n}\Big(\sum_{h=1}^{H}\alpha^{(h)}_{ij} V^{(h)}\Big)\vct{x}_j,
\qquad
\alpha^{(h)}_{ij} \;=\; \langle \phi^{(h)}(i),\psi^{(h)}(j)\rangle,
\end{equation}
and we assume causality is enforced by $\alpha^{(h)}_{ij}=0$ for $j>i$.
Assume further that $n$ is large enough that
\begin{equation}\label{eq:span-n}
\mathrm{span}\{\Wten(0),\Wten(1),\dots,\Wten(n-1)\} \;=\; \mathrm{span}\{\Wten(\tau):\tau\ge 0\}.
\end{equation}

\smallskip
\noindent\textbf{(Necessity).} If the model \eqref{eq:mh-linear-attn} represents $\mathcal{M}_{\text{SSM}}$ exactly on
length-$n$ sequences, then necessarily $H\ge k$.

\noindent\textbf{(Sufficiency).} Conversely, if $H\ge k$, then there exists a choice of parameters
$\{V^{(h)}\}_{h=1}^H$ and positional feature maps $\{\phi^{(h)},\psi^{(h)}\}_{h=1}^H$ with
\[
r \;\le\; n \qquad \text{(indeed, one may take } r = \max_{h\le k}\mathrm{rank}(A^{(h)})\le n\text{)}
\]
such that the model \eqref{eq:mh-linear-attn} represents $\mathcal{M}_{\text{SSM}}$ exactly on length-$n$ sequences.
\end{theorem}

\begin{proof}
Define the \emph{target interaction subspace}
\[
S_{\mathrm{target}} \;:=\; \mathrm{span}\{\Wten(\tau):\tau\ge 0\}\subseteq \R^{p\times d},
\qquad \dim(S_{\mathrm{target}})=k.
\]
Define the \emph{model value subspace}
\[
S_{\mathrm{MH}} \;:=\; \mathrm{span}\{V^{(1)},\dots,V^{(H)}\}\subseteq \R^{p\times d}.
\]
Clearly $\dim(S_{\mathrm{MH}})\le H$.

\par\medskip\noindent\textbf{Part I (Necessity: $H\ge k$).}
Assume the multi-head model \eqref{eq:mh-linear-attn} represents the target system exactly on length-$n$
sequences. For each pair $(i,j)$, define the (matrix-valued) kernel
\[
K_{ij}\;:=\;\sum_{h=1}^{H}\alpha^{(h)}_{ij}V^{(h)}\in\R^{p\times d}.
\]
Then $\hat{\vct{y}}_i = \sum_{j=1}^n K_{ij}\vct{x}_j$. Exact equality for \emph{all} inputs implies
equality of kernels:
\[
K_{ij} \;=\;
\begin{cases}
\Wten(i-j), & i\ge j,\\
0, & i<j,
\end{cases}
\qquad \forall\, i,j\in\{1,\dots,n\}.
\]
In particular, for each lag $\tau\in\{0,1,\dots,n-1\}$,
\[
\Wten(\tau) \;=\; K_{ij} \;=\; \sum_{h=1}^{H}\alpha^{(h)}_{ij}V^{(h)} \;\in\; S_{\mathrm{MH}}.
\]
Therefore,
\[
\mathrm{span}\{\Wten(0),\dots,\Wten(n-1)\}\;\subseteq\; S_{\mathrm{MH}}.
\]
By \eqref{eq:span-n}, $\mathrm{span}\{\Wten(0),\dots,\Wten(n-1)\}=S_{\mathrm{target}}$, hence
$S_{\mathrm{target}}\subseteq S_{\mathrm{MH}}$. Taking dimensions yields
\[
k \;=\; \dim(S_{\mathrm{target}}) \;\le\; \dim(S_{\mathrm{MH}})\;\le\; H,
\]
so necessarily $H\ge k$.

\par\medskip\noindent\textbf{Part II (Sufficiency: existence for length $n$ with $r\le n$).}
Assume $H\ge k$. Choose any basis $\{M_1,\dots,M_k\}$ of $S_{\mathrm{target}}$ and set
\[
V^{(h)} \;=\;
\begin{cases}
M_h, & 1\le h\le k,\\
0, & k<h\le H.
\end{cases}
\]
By \eqref{eq:span-n}, each $\Wten(\tau)$ for $\tau\in\{0,\dots,n-1\}$ lies in $\mathrm{span}\{M_1,\dots,M_k\}$.
Thus, for each $\tau\in\{0,\dots,n-1\}$ there exist unique scalars $c_{\tau,1},\dots,c_{\tau,k}$ such that
\begin{equation}\label{eq:W-expand}
\Wten(\tau) \;=\; \sum_{h=1}^{k} c_{\tau,h} M_h \;=\; \sum_{h=1}^{H} c_{\tau,h} V^{(h)},
\end{equation}
where we set $c_{\tau,h}=0$ for $h>k$.

For each head $h\in\{1,\dots,k\}$ define a lower-triangular coefficient matrix $A^{(h)}\in\R^{n\times n}$
by
\[
A^{(h)}_{ij} \;:=\;
\begin{cases}
c_{i-j,h}, & i\ge j,\\
0, & i<j.
\end{cases}
\]
We will choose features so that $\alpha^{(h)}_{ij}=A^{(h)}_{ij}$.

\textbf{1. Explicit factorization with $r=n$.}
Let $r=n$. For each head $h\le k$ define
\[
\phi^{(h)}(i) \;:=\; e_i\in\R^{n},\qquad
\psi^{(h)}(j) \;:=\; A^{(h)}_{:,j}\in\R^{n},
\]
where $e_i$ is the $i$th standard basis vector and $A^{(h)}_{:,j}$ is the $j$th column of $A^{(h)}$.
Then for all $i,j$,
\[
\alpha^{(h)}_{ij} \;=\; \langle \phi^{(h)}(i),\psi^{(h)}(j)\rangle \;=\; e_i^\top A^{(h)}_{:,j} \;=\; A^{(h)}_{ij}.
\]
In particular, $\alpha^{(h)}_{ij}=0$ for $i<j$, so causality holds. For heads $h>k$, set
$\phi^{(h)}\equiv 0$ (or $\psi^{(h)}\equiv 0$), yielding $\alpha^{(h)}_{ij}\equiv 0$.

Now compute the induced kernel:
for $i\ge j$,
\[
K_{ij}
\;=\;\sum_{h=1}^{H}\alpha^{(h)}_{ij}V^{(h)}
\;=\;\sum_{h=1}^{k} A^{(h)}_{ij} M_h
\;=\;\sum_{h=1}^{k} c_{i-j,h} M_h
\;=\; \Wten(i-j),
\]
where the last equality uses \eqref{eq:W-expand} with $\tau=i-j$.
For $i<j$, $K_{ij}=0$ by construction. Therefore, for all inputs,
\[
\hat{\vct{y}}_i \;=\; \sum_{j=1}^{n}K_{ij}\vct{x}_j \;=\; \sum_{j=1}^{i} \Wten(i-j)\vct{x}_j \;=\; \vct{y}_i,\qquad i=1,\dots,n,
\]
so the model represents $\mathcal{M}_{\text{SSM}}$ exactly on length-$n$ sequences.

\textbf{2. Rank-refined factorization.}
For each $h\le k$, since $A^{(h)}$ has rank $\rho_h:=\mathrm{rank}(A^{(h)})$, there exist matrices
$P^{(h)},Q^{(h)}\in\R^{n\times \rho_h}$ such that $A^{(h)}=P^{(h)}(Q^{(h)})^\top$ (e.g.\ via SVD).
Setting $\phi^{(h)}(i)$ to the $i$th row of $P^{(h)}$ and $\psi^{(h)}(j)$ to the $j$th row of $Q^{(h)}$
yields $\alpha^{(h)}_{ij}=A^{(h)}_{ij}$ with $r=\rho_h$. Taking $r=\max_{h\le k}\rho_h$ (padding with zeros)
gives $r\le n$.
\end{proof}

\smallskip
\noindent\emph{Remark.} The above sufficiency proof is constructive but may require feature dimension
$r=O(n)$, since the positional features can depend on the full length-$n$ coefficient matrices $A^{(h)}$.
Appendix~\ref{app:fixed-r} gives a stronger (more structured) construction: under mild spectral
assumptions on $\bar A$, the weights can be realized by \emph{translation-invariant} (lag-only) features,
i.e., $\alpha^{(h)}_{ij}=g_h(i-j)\,\mathbf{1}_{i\ge j}$, with feature dimension $r=O(m)$ (more generally,
$r=O(mJ^2)$), where $m$ is the state dimension and $J$ is the maximum Jordan block size of $\bar A$.



\section{Optimization Properties: Gradient Flow Analysis}
While structured dynamical models (RNNs/SSMs) possess superior expressivity (Theorems
\ref{thm:rank_gap_attn}–\ref{thm:equivalence}), they are notoriously difficult to train due to vanishing gradients
\citep{bengio1994learning, pascanu2013difficulty}. Our framework provides a rigorous characterization of this
tradeoff: expressivity comes at the cost of trainability. We formalize the gradient flow properties that make
attention mechanisms easier to optimize despite their rank limitations. Note that while recent advancements like HiPPO \citep{gu2020hippo} and S4 \citep{gu2022efficiently} mitigate this issue through specialized initialization and parameterization, the fundamental exponential decay inherent in the unconditioned dynamics remains a key theoretical distinction from the attention mechanism's topological "short-cut."

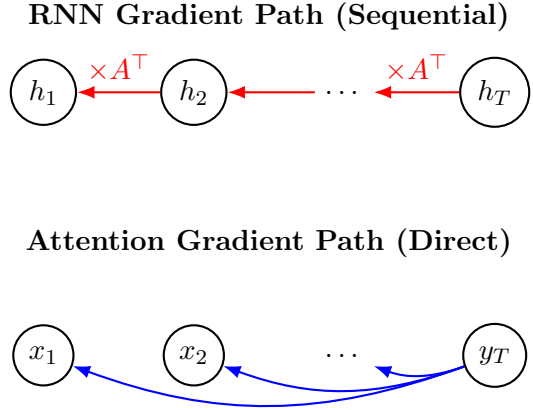
\begin{figure}[ht]
    \centering
    \begin{tikzpicture}[
        node distance=1.5cm,
        state/.style={circle, draw, minimum size=0.8cm, thick},
        arrow/.style={-Latex, thick}
    ]
        \node (rnn_label) at (0, 2) {\textbf{RNN Gradient Path (Sequential)}};
        \node[state] (h1) at (-3, 1) {$h_1$};
        \node[state] (h2) at (-1, 1) {$h_2$};
        \node (dots) at (1, 1) {$\dots$};
        \node[state] (hT) at (3, 1) {$h_T$};

        \draw[arrow, red] (hT) -- (dots) node[midway, above] {$\times A^\top$};
        \draw[arrow, red] (dots) -- (h2);
        \draw[arrow, red] (h2) -- (h1) node[midway, above] {$\times A^\top$};

        \node (attn_label) at (0, -1) {\textbf{Attention Gradient Path (Direct)}};
        \node[state] (x1) at (-3, -2.5) {$x_1$};
        \node[state] (x2) at (-1, -2.5) {$x_2$};
        \node (xdots) at (1, -2.5) {$\dots$};
        \node[state] (yT) at (3, -2.5) {$y_T$};

        \draw[arrow, blue] (yT) to[bend left=20] (x1);
        \draw[arrow, blue] (yT) to[bend left=20] (x2);
        \draw[arrow, blue] (yT) to[bend left=20] (xdots);

    \end{tikzpicture}
    \caption{\textbf{The Optimization Landscape.} Top: Recurrent models enforce a sequential gradient path of length $O(n)$, leading to exponential decay (red). Bottom: Attention mechanisms provide an $O(1)$ ``Gradient Highway'' directly to history (blue), ensuring signal preservation.}
    \label{fig:gradient_flow_viz}
\end{figure}

\begin{theorem}[Distance-Independent Gradient Paths in Attention]
\label{thm:optimization}


\leavevmode\\
Fix $1\le j < i \le n$ and define the Jacobian
\[
J_{i,j}(X)\;:=\;\frac{\partial \vct{y}_i}{\partial \vct{x}_j}\in\R^{p\times d}.
\]

\par\medskip\noindent\textbf{Part (1): Structured dynamics (linear SSM).}
Consider the linear SSM input--output map
\[
\vct{y}_i \;=\; \sum_{t=1}^{i} C\bar A^{\,i-t}\bar B\,\vct{x}_t,\qquad i=1,\dots,n.
\]
Then for every input sequence $X=(\vct{x}_1,\dots,\vct{x}_n)$,
\[
J_{i,j}(X) \;=\; C\bar A^{\,i-j}\bar B
\qquad\text{and}\qquad
\|J_{i,j}(X)\|_2 \;\le\; \|C\|_2\,\|\bar B\|_2\,\|\bar A\|_2^{\,i-j}.
\]
In particular, if $\|\bar A\|_2<1$, then $\|J_{i,j}(X)\|_2$ decays exponentially in the distance $\tau=i-j$.

\par\medskip\noindent\textbf{Part (2): Softmax attention (existence of an $O(1)$ path).}
Consider a \emph{causal} single-head softmax-attention layer
\begin{equation}\label{eq:softmax-attn-def}
\vct{y}_i \;=\; \sum_{t=1}^{i} \alpha_{it}(X)\,V\vct{x}_t,\qquad
\alpha_{it}(X)\;=\;\frac{\exp(\vct{q}_i^\top \vct{k}_t)}{\sum_{s=1}^{i}\exp(\vct{q}_i^\top \vct{k}_s)},
\qquad
\vct{q}_i=W_Q\vct{x}_i,\;\vct{k}_t=W_K\vct{x}_t,
\end{equation}
with fixed matrices $W_Q,W_K\in\R^{r\times d}$ and $V\in\R^{p\times d}$.
Assume that $\|V\|_2>0$ and that there exists $\vct{u}\in\R^d$ such that
\begin{equation}\label{eq:nondeg-assump}
\vct{a}\;:=\;W_K^\top W_Q\vct{u}\neq \vct{0}.
\end{equation}
Then for every $\varepsilon>0$ there exists an input sequence $X$ (depending on $i,j,\varepsilon$) such that
\[
\|J_{i,j}(X)\|_2 \;\ge\; \|V\|_2-\varepsilon.
\]
In particular, this lower bound is independent of the distance $\tau=i-j$.
\end{theorem}

\begin{proof}
\par\medskip\noindent\textbf{Part (1).}
Since $\vct{y}_i$ is an affine function of $(\vct{x}_1,\dots,\vct{x}_i)$ with coefficient matrices
$C\bar A^{\,i-t}\bar B$, we have
\[
\frac{\partial \vct{y}_i}{\partial \vct{x}_j}=C\bar A^{\,i-j}\bar B.
\]
The norm bound follows from submultiplicativity of $\|\cdot\|_2$:
\[
\|C\bar A^{\,i-j}\bar B\|_2 \le \|C\|_2\,\|\bar A^{\,i-j}\|_2\,\|\bar B\|_2
\le \|C\|_2\,\|\bar A\|_2^{\,i-j}\,\|\bar B\|_2.
\]

\par\medskip\noindent\textbf{Part (2).}
Fix $1\le j<i\le n$ and $\varepsilon>0$. Choose an input sequence $X=(\vct{x}_1,\dots,\vct{x}_n)$ as follows:
\[
\vct{x}_i := \vct{u},\qquad
\vct{x}_j := \frac{\gamma}{\|\vct{a}\|_2^2}\,\vct{a},\qquad
\vct{x}_t := \vct{0}\;\; \text{for all } t\in\{1,\dots,i\}\setminus\{i,j\},
\]
where $\vct{u}$ is as in \eqref{eq:nondeg-assump} and $\gamma>0$ is a scalar to be chosen later.
Then $\vct{q}_i=W_Q\vct{u}$ is fixed, $\vct{k}_j=W_K\vct{x}_j$, and the logit at position $j$ equals
\[
\vct{q}_i^\top \vct{k}_j
= (W_Q\vct{u})^\top W_K\vct{x}_j
= \vct{a}^\top \vct{x}_j
= \gamma.
\]
Moreover, for $t'\neq j$ with $1\le t'\le i$, we have $\vct{x}_{t'}=\vct{0}$ and hence $\vct{k}_{t'}=\vct{0}$,
so the corresponding logits satisfy $\vct{q}_i^\top \vct{k}_{t'}=0$.
Therefore,
\begin{equation}\label{eq:alpha-formula}
\alpha_{ij}(X)=\frac{e^{\gamma}}{e^{\gamma}+(i-1)},\qquad
1-\alpha_{ij}(X)=\frac{i-1}{e^{\gamma}+(i-1)}\le (i-1)e^{-\gamma},
\end{equation}
and
\begin{equation}\label{eq:alpha1minusalpha}
\alpha_{ij}(X)\bigl(1-\alpha_{ij}(X)\bigr)\le 1-\alpha_{ij}(X)\le (i-1)e^{-\gamma}.
\end{equation}

We now compute $J_{i,j}(X)=\partial \vct{y}_i/\partial \vct{x}_j$ for the model \eqref{eq:softmax-attn-def}.
Write $\vct{y}_i=\sum_{t=1}^{i}\alpha_{it}(X)\,V\vct{x}_t$. Differentiating w.r.t.\ $\vct{x}_j$ yields
\begin{equation}\label{eq:jacobian-split}
J_{i,j}(X)=\underbrace{\alpha_{ij}(X)\,V}_{\text{value term}}
\;+\;\underbrace{\sum_{t=1}^{i}(V\vct{x}_t)\Big(\frac{\partial \alpha_{it}}{\partial \vct{x}_j}(X)\Big)^\top}_{\text{score term (outer products)}}.
\end{equation}
Since $\vct{x}_t=\vct{0}$ for all $t\neq j$ (among $1,\dots,i$), only $t=j$ contributes to the score term, so
\begin{equation}\label{eq:jacobian-reduced}
J_{i,j}(X)=\alpha_{ij}(X)\,V \;+\; (V\vct{x}_j)\Big(\frac{\partial \alpha_{ij}}{\partial \vct{x}_j}(X)\Big)^\top.
\end{equation}

It remains to bound $\bigl\|\frac{\partial \alpha_{ij}}{\partial \vct{x}_j}(X)\bigr\|_2$.
Let $s_{it}:=\vct{q}_i^\top \vct{k}_t$ denote the logits. For softmax,
\[
\frac{\partial \alpha_{ij}}{\partial s_{ij}}=\alpha_{ij}(1-\alpha_{ij}),
\qquad
\frac{\partial s_{ij}}{\partial \vct{x}_j}
=\frac{\partial}{\partial \vct{x}_j}\bigl((W_Q\vct{x}_i)^\top (W_K\vct{x}_j)\bigr)
=W_K^\top \vct{q}_i.
\]
Therefore,
\begin{equation}\label{eq:dalpha}
\frac{\partial \alpha_{ij}}{\partial \vct{x}_j}(X)
=\alpha_{ij}(X)\bigl(1-\alpha_{ij}(X)\bigr)\,W_K^\top \vct{q}_i.
\end{equation}
In our construction, $W_K^\top \vct{q}_i=W_K^\top W_Q\vct{u}=\vct{a}$, hence
\begin{equation}\label{eq:dalpha-norm}
\Bigl\|\frac{\partial \alpha_{ij}}{\partial \vct{x}_j}(X)\Bigr\|_2
=\alpha_{ij}(X)\bigl(1-\alpha_{ij}(X)\bigr)\,\|\vct{a}\|_2.
\end{equation}

Now bound the perturbation term in \eqref{eq:jacobian-reduced}:
\[
\Bigl\|(V\vct{x}_j)\Big(\frac{\partial \alpha_{ij}}{\partial \vct{x}_j}(X)\Big)^\top\Bigr\|_2
\;\le\; \|V\vct{x}_j\|_2\;\Bigl\|\frac{\partial \alpha_{ij}}{\partial \vct{x}_j}(X)\Bigr\|_2.
\]
Using $\|V\vct{x}_j\|_2\le \|V\|_2\|\vct{x}_j\|_2$ and $\|\vct{x}_j\|_2=\gamma/\|\vct{a}\|_2$, together with
\eqref{eq:dalpha-norm} and \eqref{eq:alpha1minusalpha}, we obtain
\begin{equation}\label{eq:perturb-bound}
\Bigl\|(V\vct{x}_j)\Big(\frac{\partial \alpha_{ij}}{\partial \vct{x}_j}(X)\Big)^\top\Bigr\|_2
\;\le\; \|V\|_2\cdot \frac{\gamma}{\|\vct{a}\|_2}\cdot \alpha_{ij}(1-\alpha_{ij})\|\vct{a}\|_2
\;\le\; \|V\|_2\,\gamma\,(i-1)e^{-\gamma}.
\end{equation}

Finally, by the reverse triangle inequality and \eqref{eq:jacobian-reduced},
\[
\|J_{i,j}(X)\|_2
\;\ge\; \|\alpha_{ij}(X)V\|_2 - \Bigl\|(V\vct{x}_j)\Big(\frac{\partial \alpha_{ij}}{\partial \vct{x}_j}(X)\Big)^\top\Bigr\|_2
\;=\; \alpha_{ij}(X)\|V\|_2 - \Bigl\|(V\vct{x}_j)\Big(\frac{\partial \alpha_{ij}}{\partial \vct{x}_j}(X)\Big)^\top\Bigr\|_2.
\]
Choose $\gamma$ large enough so that simultaneously
\[
1-\alpha_{ij}(X)\le \frac{\varepsilon}{2\|V\|_2}
\qquad\text{and}\qquad
\|V\|_2\,\gamma\,(i-1)e^{-\gamma}\le \frac{\varepsilon}{2}.
\]
This is always possible because $1-\alpha_{ij}(X)\le (i-1)e^{-\gamma}$ by \eqref{eq:alpha-formula}
and $\gamma e^{-\gamma}\to 0$ as $\gamma\to\infty$.
With this choice,
\[
\|J_{i,j}(X)\|_2
\;\ge\; \Bigl(1-\frac{\varepsilon}{2\|V\|_2}\Bigr)\|V\|_2 - \frac{\varepsilon}{2}
\;=\; \|V\|_2-\varepsilon,
\]
which proves the claim.
\end{proof}

\begin{remark}[Capacity vs. Guarantee]
Theorem \ref{thm:optimization} establishes the existence of a sequence $X$ that maintains a large gradient norm,
thereby proving that the attention architecture possesses the topological \textit{capacity} to propagate signals
over arbitrary distances without attenuation. This is distinct from providing a \textit{guarantee} that gradients
will be stable for random initializations or average-case data, although it highlights the structural advantage
over architectures where no such path exists.
\end{remark}

\section{Numerical Verification}
\label{sec:experiments}

To complement our theoretical results, we perform controlled synthetic experiments probing (i) the head-count
expressivity constraint predicted by Theorem~\ref{thm:equivalence} and (ii) the qualitative gradient-flow
behavior highlighted by Theorem~\ref{thm:optimization}.

\subsection{Experiment I: The Interaction Rank Gap}
We study a system identification task to empirically probe Theorem~\ref{thm:equivalence}.
\textbf{Setup:} We define a ``teacher'' linear SSM with state dimension $k$. The transition matrix $\bar{A}$ is
constructed from block-diagonal rotation matrices so that the induced impulse-response family has interaction rank $k$.
A ``student'' multi-head factorized linear-attention model with $H$ heads is trained to match the teacher's outputs.
To isolate structural expressivity, the student uses learnable positional encodings (to fit the lag structure) and a
linear-attention form consistent with the theorem's setting. We vary the target rank $k\in\{2,4,8\}$ and the number of
student heads $H\in\{1,\dots,9\}$.

\begin{figure}[ht]
    \centering
    \IfFileExists{experiment_results_full.png}{
        \includegraphics[width=1\textwidth]{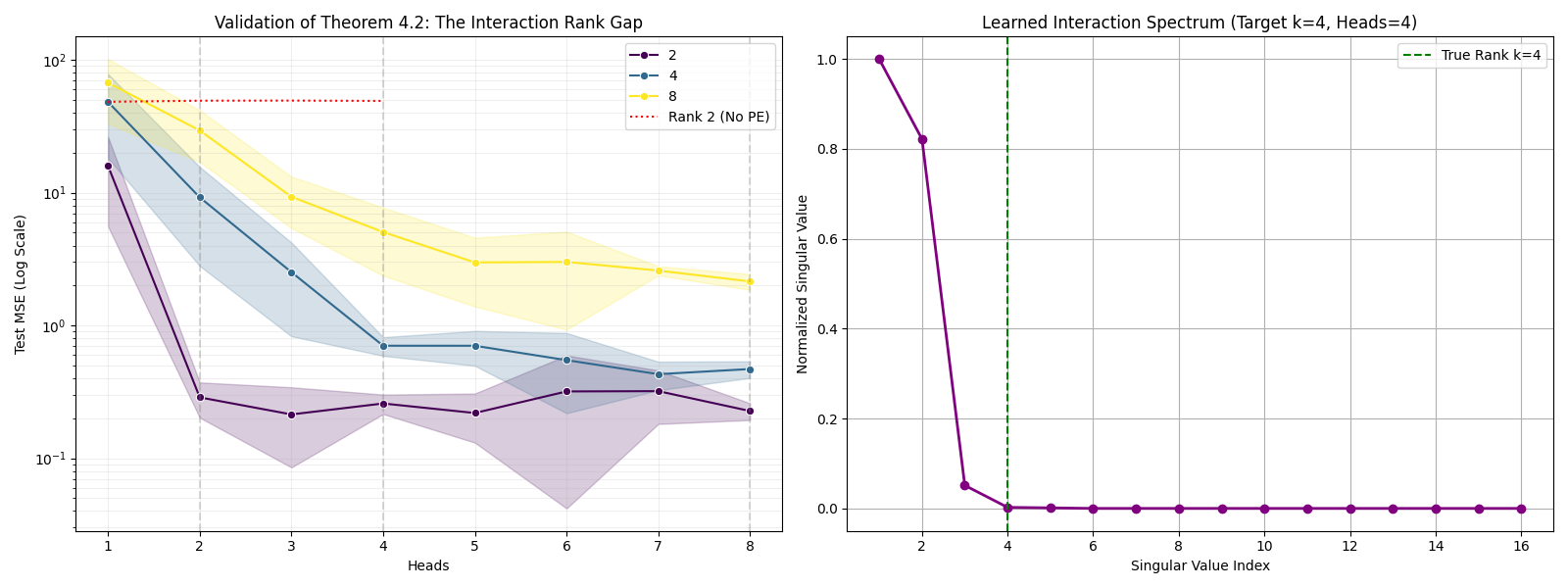}
    }{
        \framebox{\parbox{1\textwidth}{\centering
            \vspace{3cm}
            \textbf{Image File Not Found} \\
            \small\textit{experiment\_results\_full.png}
            \vspace{3cm}
        }}
    }
    \caption{Left: Test MSE vs.\ number of heads for different target interaction ranks $k$. Error drops sharply as
    $H$ approaches $k$ and reaches a low-error regime once $H\ge k$, consistent with Theorem~\ref{thm:equivalence}.
    Right: Singular-value spectrum of the learned interaction operator for $k=4$, $H=4$, showing $\approx 4$
    dominant components.}
    \label{fig:rank_gap}
\end{figure}

\textbf{Results:} Figure~\ref{fig:rank_gap} (Left) reports the test MSE after training. We observe a pronounced
error drop as $H$ approaches the target interaction rank $k$, with consistently low error once $H\ge k$.
For example, for $k=8$ (blue curve), performance improves steadily with $H$ and only reaches the low-error regime
once $H$ is sufficiently large. This behavior is consistent with the necessity direction of
Theorem~\ref{thm:equivalence}: when $H<k$, the model is restricted to a value subspace of dimension at most $H$ and
cannot, in general, represent a $k$-dimensional interaction family. Moreover, for $H<k$ the monotone improvement
suggests that additional heads capture progressively larger subspaces of the underlying dynamics.

\textbf{Mechanism verification:} Figure~\ref{fig:rank_gap} (Right) shows the singular-value spectrum of the learned
interaction operator for a $k=4$ teacher trained with $H=4$ heads. The spectrum exhibits approximately $4$ dominant
components (with the remainder near the numerical floor), consistent with the interpretation that the $H$ heads learn
a basis spanning the target $k$-dimensional interaction subspace.

\subsection{Experiment II: Gradient Flow Dynamics}
\label{sec:exp_gradient}
We next examine gradient-flow behavior in the spirit of Theorem~\ref{thm:optimization}.
\textbf{Setup:} We compare a linear SSM and a multi-head attention model under standard initialization. For each
sequence length $T$, we measure the norm of the sensitivity of a late output to the first input token,
$\|\nabla_{x_0} y_T\|$, as a function of $T$ (semi-log scale).

\textbf{Prediction:} Theorem~\ref{thm:optimization} implies that for a stable linear SSM the input--output Jacobian
decays with distance (and can decay exponentially under spectral contraction). In contrast, attention admits
distance-independent gradient paths for suitable inputs; in practice, this suggests substantially slower decay of
long-range sensitivities compared to a stable linear SSM.

\begin{figure}[ht]
    \centering
    \IfFileExists{experiment_gradient_flow.png}{
        \includegraphics[width=0.8\textwidth]{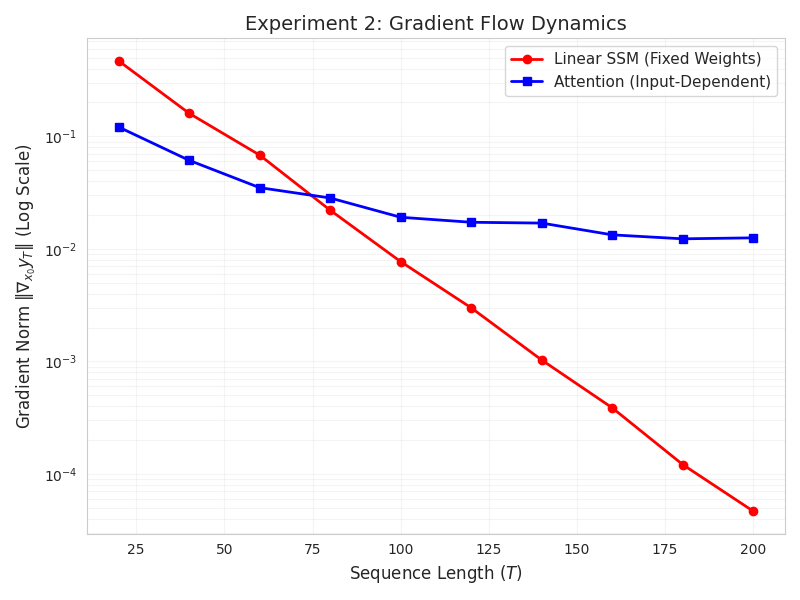}
    }{
        \framebox{\parbox{0.8\textwidth}{\centering
            \vspace{3cm}
            \textbf{Image File Not Found} \\
            \small\textit{experiment\_gradient\_flow.png}
            \vspace{3cm}
        }}
    }
    \caption{Gradient norm $\|\nabla_{x_0} y_T\|$ vs.\ sequence length $T$ (semi-log scale). The linear SSM decays
    exponentially, while attention decays much more slowly and remains larger over long contexts, consistent with
    Theorem~\ref{thm:optimization}.}
    \label{fig:gradient_flow}
\end{figure}

\textbf{Results:} Figure~\ref{fig:gradient_flow} shows the measured gradient norms. The linear SSM (red) exhibits a
clear exponential decay with $T$, consistent with the submultiplicative bound in Theorem~\ref{thm:optimization}. The
attention model (blue) decays much more slowly and remains substantially larger across the tested lengths, illustrating
the mechanism that attention can preserve long-range sensitivity relative to stable linear dynamics.

\section{Discussion: Bridging the Gap with Hybrid Architectures}
\label{sec:discussion}
Our analysis highlights a tension in sequence modeling. The \textbf{Interaction Rank Gap}
(Theorem~\ref{thm:rank_gap_attn}) shows that matching a $k$-dimensional interaction family requires at least $k$ heads
in a multi-head factorized attention model, while the \textbf{Gradient Sensitivity Bound}
(Theorem~\ref{thm:optimization}) emphasizes that stable linear dynamics can attenuate long-range sensitivities.

This perspective helps rationalize \textbf{hybrid architectures} that combine structured SSM updates with occasional
attention layers. A prominent example is \textbf{Jamba} \citep{lieber2024jamba}, which interleaves Mamba (SSM) blocks
with Transformer (attention) layers. Under our framework:
\begin{itemize}
    \item \textbf{SSM layers} efficiently implement rich structured dynamics and can capture high-rank state evolution
    without relying on attention across all positions.
    \item Periodic \textbf{attention layers} can provide direct long-range interaction and help preserve gradient
    signal over long contexts by introducing non-local pathways, mitigating the distance-dependent attenuation present
    in stable linear updates.
\end{itemize}
In this view, Jamba-style hybrids can be interpreted as an engineering response to the rank--trainability trade-off
formalized by our results.

\section{Conclusion}
\label{sec:conclusion}
We presented a Unified  Framework that places prominent sequence modeling paradigms within a common mathematical
structure. By contrasting scalar-factorized interactions (attention) with structured dynamics (SSMs), we derived
rigorous constraints on expressivity (via interaction rank) and on long-range sensitivity (via Jacobian bounds). Our
results show that multi-head attention is not merely an ensemble heuristic: head count controls the dimension of the
interaction subspace it can represent. Conversely, attention mechanisms can sustain stronger long-range sensitivities
than stable linear dynamics by creating non-local pathways. Together, these insights provide a principled lens for
designing architectures that balance structured dynamical modeling with trainability over long contexts.


\appendix
\section{Fixed-Dimensional Translation-Invariant Features under Spectral Assumptions}\label{app:fixed-r}

In Theorem~\ref{thm:equivalence}, the sufficiency construction guarantees exact representability on
length-$n$ sequences with feature dimension $r\le n$. In this appendix we record a common strengthening:
under additional assumptions on $\bar A$, one can realize the coefficients $c_{\tau,h}$ with
\emph{translation-invariant} features of dimension $r=O(m)$, where $m$ is the state dimension.

\paragraph{Standing assumptions.}
Let $\bar A\in\R^{m\times m}$ be the state transition matrix.
Assume:
\begin{enumerate}[label=(A\arabic*)]
\item \textbf{Bounded Jordan degree:} Over $\mathbb{C}$, every Jordan block of $\bar A$ has size at most $J$
(for diagonalizable $\bar A$, $J=1$).
\item \textbf{No zero eigenvalues (invertibility):} $\bar A$ is invertible (equivalently, $0$ is not an eigenvalue).
\end{enumerate}
Assumption (A2) is used only to express $\lambda^{i-j}$ as $\lambda^i\lambda^{-j}$ with finite features; if
$\bar A$ is singular, additional bookkeeping is required (e.g., splitting the nilpotent part), and we omit it here.

\paragraph{Connection to common SSM parameterizations.}
Assumptions (A1)--(A2) are broadly compatible with several structured, time-invariant SSM layers used in
practice. For example, the S4 family parameterizes the transition so that powers of $\bar A$ (and hence the
convolution kernel) admit efficient representations based on spectral structure \cite{gu2022efficiently},
and related constructions originate from the HiPPO framework \cite{gu2020hippo}. In the special case where
$\bar A$ is diagonalizable (so $J=1$ in (A1)), the translation-invariant feature factorization below becomes
particularly simple. Assumption (A2) is mainly a technical convenience for writing $\lambda^{i-j}=\lambda^i\lambda^{-j}$;
many discretizations used in SSM sequence models yield invertible discrete-time transitions, and singular cases can be
handled with additional bookkeeping (as noted above). Finally, we emphasize that the translation-invariant features in
this appendix apply to \emph{fixed} (input-independent) kernels $W(\tau)$; selective/state-dependent variants such as
Mamba depart from strict translation invariance and require separate analysis \cite{gu2023mamba}.


\paragraph{Mode decomposition.}
Let $\{\lambda_\ell\}_{\ell=1}^{L}\subset\mathbb{C}$ denote the distinct eigenvalues of $\bar A$ (counted without
multiplicity). Under (A1), by the standard Jordan normal form power identity, $\bar A^\tau$ is a finite sum of terms of the form $\lambda_\ell^\tau p(\tau)$. Rigorously, there exist matrices $G_{\ell,q}\in\mathbb{C}^{m\times m}$ and polynomials $p_{\ell,q}(\tau)$ such that
\begin{equation}\label{eq:jordan-modes}
\bar A^\tau \;=\; \sum_{\ell=1}^{L}\;\sum_{q=0}^{J_\ell-1} \lambda_\ell^\tau\,p_{\ell,q}(\tau)\,G_{\ell,q}.
\end{equation}
Since polynomials can be expressed in the monomial basis $\{\tau^q\}$, we can absorb coefficients to rewrite this as a sum over terms $\lambda_\ell^\tau \tau^q$ (potentially redefining the matrices $G_{\ell,q}$ or introducing coefficients). Multiplying by $C$ on the left and $\bar B$ on the right yields a similar expansion for $\Wten(\tau)$:
\begin{equation}\label{eq:W-modes}
\Wten(\tau)=C\bar A^\tau \bar B
\;=\;\sum_{\ell=1}^{L}\;\sum_{q=0}^{J_\ell-1} \lambda_\ell^\tau\,\tau^{q}\,H_{\ell,q},
\qquad H_{\ell,q} \in \mathbb{C}^{p\times d}.
\end{equation}
Consequently, for any fixed basis $\{M_h\}_{h=1}^k$ of $S_{\mathrm{target}}$ and dual basis linear functionals
$\{\mathcal{L}_h\}_{h=1}^k$ (i.e., $\mathcal{L}_h(M_{h'})=\delta_{hh'}$), the coefficient sequences
$c_{\tau,h}=\mathcal{L}_h(\Wten(\tau))$ satisfy
\begin{equation}\label{eq:c-modes}
c_{\tau,h}
\;=\;\sum_{\ell=1}^{L}\;\sum_{q=0}^{J_\ell-1} b_{h,\ell,q}\,\lambda_\ell^\tau\,\tau^{q},
\end{equation}
for some scalars $b_{h,\ell,q}\in\mathbb{C}$.

\paragraph{Translation-invariant feature factorization.}
Fix a head $h$. Define the lag kernel $g_h(\tau):=c_{\tau,h}$ for $\tau\ge 0$ (and implicitly whatever value the polynomial gives for $\tau < 0$, though we will mask it).
Under (A2), each mode $\lambda_\ell^{i-j}(i-j)^q$ can be factorized as an inner product of features of $i$ and $j$
by expanding $(i-j)^q=\sum_{t=0}^{q}\binom{q}{t} i^{t}(-j)^{q-t}$ and using $\lambda_\ell^{i-j}=\lambda_\ell^i\lambda_\ell^{-j}$:
\begin{equation}\label{eq:mode-factor}
\lambda_\ell^{i-j}(i-j)^q
=
\sum_{t=0}^{q}\binom{q}{t}\big(i^{t}\lambda_\ell^i\big)\big((-j)^{q-t}\lambda_\ell^{-j}\big).
\end{equation}
Thus, defining (complex-valued) feature vectors indexed by $(\ell,q,t)$,
\[
\phi^{(h)}(i)_{(\ell,q,t)} := i^{t}\lambda_\ell^{i},
\qquad
\psi^{(h)}(j)_{(\ell,q,t)} := \binom{q}{t}\,b_{h,\ell,q}\,(-j)^{q-t}\lambda_\ell^{-j},
\]
yields $\langle \phi^{(h)}(i),\psi^{(h)}(j)\rangle = g_h(i-j)$ for all $i,j$, where $g_h$ is the extension of the formula in \eqref{eq:c-modes} to $\mathbb{Z}$.
However, the attention model requires $\alpha_{ij} = 0$ for $i < j$. Therefore, we explicitly define the weights with a causal mask:
\[ \alpha^{(h)}_{ij} = g_h(i-j)\cdot \mathbf{1}_{i \ge j}. \]
This ensures causality.
The total feature dimension per head is
\[
r_h \;\le\; \sum_{\ell=1}^{L}\sum_{q=0}^{J_\ell-1}(q+1)\;\le\;\sum_{\ell=1}^{L}\frac{J_\ell(J_\ell+1)}{2}
\;\le\; \frac{LJ(J+1)}{2}
\;\le\; O(mJ^2).
\]
Since $L\le m$ and $J_\ell\le J$. This bound becomes $O(m)$ when the maximum Jordan block size $J$ is treated as a constant (e.g., $J=1$ for diagonalizable matrices). Converting complex conjugate pairs to real features yields an equivalent real construction.

\paragraph{Conclusion.}
Under assumptions (A1)--(A2), the coefficient sequences $c_{\tau,h}$ admit translation-invariant inner-product
representations with feature dimension $r=O(mJ^2)$ per head (or $O(m)$ for constant $J$). Plugging these $\alpha^{(h)}_{ij}=g_h(i-j)\mathbf{1}_{i \ge j}$ into the
multi-head decomposition $\sum_{h=1}^k \alpha^{(h)}_{ij}V^{(h)}$ recovers the same basis expansion used in the
proof of Theorem~\ref{thm:equivalence}.

\bibliographystyle{plainnat}
\bibliography{references}

\end{document}